\pdfoutput=1

\documentclass[11pt]{article}
\usepackage[]{acl}

\usepackage{times}
\usepackage{latexsym}
\usepackage[T1]{fontenc}
\usepackage[utf8]{inputenc}
\usepackage{microtype}

\newcommand{\eg}{\textit{e.g.} }

\newcommand{\semeval}{SemEval }
\newcommand{\wordnet}{WordNet}
\newcommand{\babelnet}{BabelNet}
\newcommand{\glove}{GloVe}

\newcommand{\wic}{WiC}
\newcommand{\scalablewsi}{\textit{Scalable WSI}}
\newcommand{\ourds}{\textit{25-7-1-8}}

\usepackage{microtype}
\usepackage{graphicx}
\usepackage{amsmath}
\usepackage{multirow}
\usepackage{tcolorbox}

\usepackage{siunitx}
\usepackage{tikz}
\usepackage{pgfplots}
\pgfplotsset{width=7cm,compat=1.3}

\title{Large Scale Substitution-based Word Sense Induction}

\author{Matan Eyal\textsuperscript{1} \quad Shoval Sadde\textsuperscript{1} \quad Hillel Taub-Tabib\textsuperscript{1} \quad Yoav Goldberg\textsuperscript{1,2}  \\
  \textsuperscript{1} Allen Institute for AI, Israel \\
  \textsuperscript{2} Bar Ilan University, Ramat-Gan, Israel \\
  \texttt{{matane,shovals,hillelt,yoavg}@allenai.org}
  }

\begin{document}
\maketitle
\begin{abstract}
We present a word-sense induction method based on pre-trained masked language models (MLMs), which can cheaply scale to large vocabularies and large corpora.
The result is a corpus which is sense-tagged according to a corpus-derived sense inventory and where each sense is associated with indicative words.
Evaluation on English Wikipedia that was sense-tagged using our method shows that both the induced senses, and the per-instance sense assignment, are of high quality even compared to WSD methods, such as Babelfy.
Furthermore, by training a static word embeddings algorithm on the sense-tagged corpus, we obtain high-quality static senseful embeddings. These outperform existing senseful embeddings methods on the WiC dataset and on a new outlier detection dataset we developed.
The data driven nature of the algorithm allows to induce corpora-specific senses, which may not appear in standard sense inventories, as we demonstrate using a case study on the scientific domain.
\end{abstract}

\begin{figure*}
\small
\resizebox{0.99\linewidth}{!}{%
\begin{tabular}{lllll|lllll}
\multicolumn{10}{l}{\large\color{purple} bug}\\
\multicolumn{5}{l}{\color{gray}Representatives} & \multicolumn{5}{l}{\color{gray}Neighbours}\\
\textbf{bug$_0$} & \textbf{bug$_1$} & \textbf{bug$_2$} & \textbf{bug$_3$} & \textbf{bug$_4$} & \textbf{bug$_0$} & \textbf{bug$_1$} & \textbf{bug$_2$} & \textbf{bug$_3$} & \textbf{bug$_4$} \\
insect & problem & feature & bomb & virus & bugs$_0$ & vulnerability$_2$ & bugs$_1$ & bugs$_3$ & flu$_2$ \\
fly & flaws & fix & device & infection & beetle$_0$ & glitch & patches$_2$ & dumpster & staph \\
beetle & hole & code & bite & crisis & spider$_0$ & rootkit & bug$_1$ & laptop$_1$ & hangover \\
Bugs & patch & dog & screen & disease & snake$_1$ & bugs$_1$ & updates$_1$ & footage$_1$ & nosebleed \\
worm & mistake & software & tag & surprise & worm$_0$ & virus$_2$ & patch$_2$ & cruiser$_3$ & pain$_4$ \\
\end{tabular}
}

\noindent\makebox[\linewidth]{\rule{\linewidth}{0.4pt}}

\resizebox{0.95\linewidth}{!}{%
\begin{tabular}{ll|ll||ll|ll}
\multicolumn{4}{l||}{\large\color{purple} Java} & \multicolumn{4}{l}{\large\color{purple} chair}\\
\multicolumn{2}{l}{\color{gray}Representatives} & \multicolumn{2}{l||}{\color{gray}Neighbours} & \multicolumn{2}{l}{\color{gray}Representatives} & \multicolumn{2}{l}{\color{gray}Neighbours}\\
\textbf{Java$_0$} & \textbf{Java$_1$} & \textbf{Java$_0$} & \textbf{Java$_1$} & \textbf{chair$_0$} & \textbf{chair$_1$} & \textbf{chair$_0$} & \textbf{chair$_1$} \\
Jakarta & Eclipse & Timor$_0$ & Python$_0$ & head & seat & Chair$_0$ & stool$_0$ \\
Indonesia & Jo & Sumatra$_1$ & JavaScript & chairman & position & chairperson & podium$_2$ \\
Bali & Apache & Sulawesi & Pascal$_2$ & president & wheelchair & chairman$_0$ & desk$_0$ \\
Indies & software & Sumatra$_0$ & SQL & presided & professor & president$_0$ & professorship \\
Holland & Ruby & Kalimantan & library$_3$ & lead & table & Chairman$_0$ & throne$_1$
\end{tabular}
}

\smallskip

\noindent\makebox[\linewidth]{\rule{\linewidth}{0.4pt}}

\resizebox{0.99\linewidth}{!}{%
\begin{tabular}{lll|lll||ll|ll}
\multicolumn{6}{l||}{\large\color{purple} pound} & \multicolumn{4}{l}{\large\color{purple} train}\\
\multicolumn{3}{l}{\color{gray}Representatives} & \multicolumn{3}{l||}{\color{gray}Neighbours} & \multicolumn{2}{l}{\color{gray}Representatives} & \multicolumn{2}{l}{\color{gray}Neighbours}\\
\textbf{pound$_0$} & \textbf{pound$_1$} & \textbf{pound$_2$} & \textbf{pound$_0$} &
\textbf{pound$_1$} & \textbf{pound$_2$} &\textbf{train$_0$} & \textbf{train$_1$} & \textbf{train$_0$} & \textbf{train$_1$} \\
lb & dollar & beat & lb$_0$ & rupee & smash$_2$ & training & railway & recruit$_0$ & bus$_0$ \\
foot & marks & punch & pounds$_0$ & shilling & kick$_1$ & prepare & track & equip & tram$_1$ \\
weight & coin & pump & lbs$_0$ & dollar$_1$ & stomp & educate & rail & recruit$_1$ & trains$_1$ \\
ton & Mark & crush & ton$_2$ & franc & slash$_0$ & practice & line & volunteer$_2$ & carriage$_0$ \\
kilograms & mile & attack & lbs$_1$ & penny$_0$ & throw$_4$ & qualified & railroad & retrain & coach$_3$ \\
\end{tabular}
}
\caption{Examples of induced word-senses for various words.  For each sense we list the top-5 representatives, as well as the 5 closest neighbours in the static embeddings space.}
\vspace{-1.5em}
\label{fig:examples}
\end{figure*}

\section{Introduction}
Word forms are ambiguous, and derive meaning from the context in which they appear.
For example, the form ``bass'' can refer to a musical instrument, a low-frequency sound, a type of voice, or a kind of fish. The correct reference is determined by the surrounding linguistic context.
Traditionally, this kind of ambiguity was dealt via \emph{word sense disambiguation} (WSD),
a task that disambiguates word forms in context between symbolic sense-ids from a sense inventory such as WordNet \cite{miller1995wordnet} or, more recently, BabelNet \cite{navigli-ponzetto-2010-babelnet}. Such sense inventories rely heavily on manual curation, are labor intensive to produce, are not available in specialized domains and inherently unsuitable for words with emerging senses.\footnote{For example, in current \wordnet~version, \textit{Corona} has 6 synsets, none of them relates to the novel \textit{Coronavirus}.}
This can be remedied by \emph{word sense induction} (WSI), a task where the input is a given word-type and a corpus, and the output is a derived sense inventory for that word. Then, sense disambiguation can be performed over the WSI-derived senses.

The introduction of large-scale pre-trained LMs and Masked LMs (MLM) seemingly made WSI/WSD tasks obsolete: instead of representing tokens with symbols that encode sense information, each token is associated with a contextualized vector embeddings that captures various aspects of its in-context semantics, including the word-sense. These contextualized vectors proved to be very effective as features for downstream NLP tasks. However, contextualized embeddings also have some major shortcomings: most notably for our case, they are expensive to store (\eg BERT embeddings are 768 or 1024 floating point numbers for each token), and are hard to index and query at scale. Even if we do manage to store and query them, they are not interpretable, making it impossible for a user to query for a particular sense of a word without providing a full disambiguating context for that word. For example, consider a user wishing to query a dataset for sentences discussing \emph{Oracle} in the mythology-prophet sense, rather than the tech company sense. It is not clear how to formulate such a query to an index of contextualized word vectors. However, it is trivial to do for an index that annotates each token with its derived sense-id (in terms of UI, after a user issues a query such as ``Oracle'', the system may show a prompt such as ``did you mean Oracle related to IBM; Sun; Microsoft, or to Prophet; Temple; Queen'', allowing to narrow the search in the right direction).

\citet{amrami2018word, amrami2019towards} show how contextualized embeddings can be used for achieving state-of-the-art WSI results. The core idea of their WSI algorithm is based on the intuition, first proposed by \citet{bacskaya2013ai}, that occurrences of a word that share a sense, also share in-context substitutes. An MLM is then used to derive top-$k$ word substitutes for each word, and these \emph{substitute-vectors} are clustered to derive word senses.

Our main contribution in this work is proposing a method that scales up \citet{amrami2018word}'s work to \emph{efficiently} annotate all tokens in a large corpus (e.g. Wikipedia) with automatically derived word-senses. This combines the high-accuracy of the MLM-based approach, with the symbolic representation provided by discrete sense annotations. The discrete annotations are interpretable (each sense is represented as a set of words), editable, indexable and searchable using standard IR techniques. 
We show two applications of the discrete annotations, the first one is sense-aware information retrieval (\S\ref{section:senseful_ir}), and the second is high-quality senseful \emph{static} word embeddings we can derive by training a static embeddings model on the large sense annotated corpus (\S\ref{section:senseful_w2v}).

We first show how the method proposed by \citet{amrami2018word} can be adapted from deriving senses of individual lemmas to efficiently and cheaply annotating \emph{all the corpus occurrences} of \emph{all the words in a large vocabulary} (\S\ref{section:wsi_at_scale}). Deriving word-sense clusters for all of English Wikipedia words that appear as single-token words in \textsc{BERTlarge}'s \cite{devlin-etal-2019-bert} vocabulary, and assigning a sense to each occurrence in the corpus, required 100 hours of cheap P100 GPUs (5 hours of wall-clock time on 20 single GPU machines) followed by roughly 4 hours on a single 96-cores CPU machines. The whole process requires less than 50GB of disk space, and costs less than 150\$ on Google Cloud platform.

After describing the clustering algorithm (\S\ref{subsection:community_detection}), we evaluate the quality of our system and of the automatic sense tagging using \semeval datasets and a new manually annotated dataset we created (\S\ref{section:evaluation}). We show that with the produced annotated corpora it is easy to serve sense-aware information retrieval applications (\S\ref{section:senseful_ir}).
Another immediate application is feeding the sense-annotated corpora to a static embedding algorithm such as word2vec \cite{mikolov2013distributed}, for deriving \emph{sense-aware static embeddings} (\S\ref{section:senseful_w2v}). This results in state-of-the-art sense-aware embeddings, which we evaluate both on an existing WiC benchmark \cite{pilehvar-camacho-collados-2019-wic} and on a new challenging benchmark which we create (\S\ref{section:embs_eval}).

In contrast to WSD which relies on curated sense inventories, our method is data-driven, therefore resulting senses are corpus dependent. The method can be applied to any domain for which a BERT-like model is available, as we demonstrate by applying it to the PubMed Abstracts of scientific papers, using \textsc{SciBERT} \cite{Beltagy2019SciBERT}. The resulting senses cover scientific terms which are not typically found in standard sense inventories (\S\ref{section:biomedical_data}).

Figure \ref{fig:examples} shows examples of induced senses for selected words from the English Wikipedia corpus. For each sense we list 5 community-based representatives (\S\ref{section:wsi_at_scale}), as well as the 5 closest neighbours in the sense-aware embedding space (\S\ref{section:senseful_w2v}). Additional examples are available in Appendix \ref{appendix:more_examples}. Code and resources are available in \href{https://github.com/allenai/WSIatScale}{github.com/allenai/WSIatScale}.

\section{Related Work}
\label{section:related_work}

\paragraph{Word Sense Induction and Disambiguation}
Previous challenges like \citet{jurgens2013semeval} focused on word sense induction for small sized datasets. To the best of our knowledge we are the first to perform large-scale \textit{all-words} WSI. The closest work to our method is the substitution-based method proposed in \citet{amrami2018word, amrami2019towards} which is the starting point to our paper. In that paper, the authors suggested a WSI algorithm designed for a small dataset (\semeval 2010, 2013) with a predefined set of ambiguous target words (See (\S\ref{section:wsi_at_scale}) for more details on the algorithm). In our work, we change \citet{amrami2019towards} such that we can efficiently run sense induction on all the words in very large corpora.

An alternative approach for sense tagging is based on Word Sense Disambiguation (WSD). The two main WSD methods are Supervised WSD and Knowledge-based WSD. Supervised WSD suffers from the difficulty of
obtaining an adequate amount of annotated data. Indeed, even SemCor, the largest manually annotated tagged corpus, consists of only 226,036 annotated tokens.
Among different supervisied WSD methods, \citet{zhong-ng-2010-makes} suggested a SVM based approach and \citet{melamud-etal-2016-context2vec, yuan2016semi} suggested LSTMs paired with nearest neighbours classification.
Knowledge-base WSD \cite{moro2014entity, pasini2017train}, on the other hand, avoids the reliance on large annotated word-to-sense corpus and instead maps words to senses from a closed sense inventory (\eg \wordnet~\cite{miller1995wordnet}, \babelnet~\cite{navigli-ponzetto-2010-babelnet}). As such, the quality of knowledge-based WSD heavily depends on the availability, quality and coverage of the associated annotated resources. 
\paragraph{Sense Embeddings}
In \S\ref{section:senseful_w2v} we exploit the sense-induced corpus to train sense embeddings. \citet{reisinger2010multi} were the first to suggest creating multiple representations for ambiguous words. Numerous recent papers \cite{chen2014unified, rothe2015autoextend, iacobacci-etal-2015-sensembed, pilehvar2016conflated, mancini-etal-2017-embedding, iacobacci2019lstmembed} aim to produce similar embeddings, all of which use either \wordnet~or \babelnet~as semantic network. Our method is similar to \citet{iacobacci-etal-2015-sensembed}, with the difference being that they rely on semantic networks (via Babelfy \cite{moro2014entity}). In contrast and similarly to us, \citet{pelevina2017making} does not rely on lexical resources such as \wordnet. 
The authors proposed splitting pretrained embeddings (such as word2vec) to a number of prototype sense-embeddings. Yet in our work, we directly learn the multi-prototype sense-embeddings which is only possible due to the large-scale corpus annotation. When comparing both methods in \S\ref{section:wic} we infer it is better to directly learn multi-prototype sense-embeddings.

\begin{figure*}[t!]
\centering
\includegraphics[width=\linewidth]{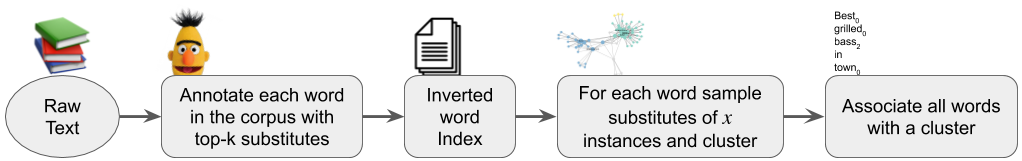}
\caption{\scalablewsi~flow. Given raw text, we annotate each word with its top-k substitutes, create inverted word index, find best clusters for each distinct lemma and associate all corpus words with a matching cluster.}
\label{fig:flow}
\vspace{-4mm}
\end{figure*}

\section{Large Scale Sense Induction}
\label{section:wsi_at_scale}

\subsection{Definition}
We define large-scale sense induction as deriving sense clusters for all words in a large vocabulary and assigning a sense cluster to each corpus occurrence of these words.\footnote{
In \textit{BERT-large-cased-whole-word-masking} this corresponds to 16k vocabulary items, that match to 1.59B full words in English Wikipedia, or 92\% of all word occurrences. Analyzing the remaining words, only 0.01\% appear in Wikipedia more than 100 times.
We derive word senses to a substantial chunk of the vocabulary, which also corresponds to the most ambiguous words as less frequent words are substantially less polysemous \cite{hernandez2016testing, fenk2010frequency, zipf1945meaning}.
}

\subsection{Algorithm}
Contextualized BERT vectors contain sense information, and clustering the contextualized vectors results in sense clusters. However, storing a 1024 dimensional vector of 32bit floats for each relevant token in the English Wikipedia corpus requires over 8TB of disk-space, making the approach cumbersome and not-scalable. However, as shown by \citet{amrami2019towards}, MLM based word-substitutes also contain the relevant semantic information, and are much cheaper to store: each word-id in \textsc{BERTlarge}'s vocabulary can be represented by 2 bytes, and storing the top-5 substitutes for each corpus position requires less than 20GB of storage space.\footnote{The size can be reduced further using adaptive encoding techniques that assign fewer bits to frequent words. We did not implement this in this work.}

In order to perform WSI at scale, we keep the main intuition from \citet{amrami2019towards}, namely to cluster sparse vectors of lemmas of the top-k MLM-derived word substitutions. This results in vast storage saving, and also in a more interpretable representations. However, for scalability, we iterate over the corpus sentences and collect the top-k substitutes for all words in the sentence at once based on a single BERT call for that sentence. This precludes us from using the dynamic-patterns component of their method, which requires separately running BERT for each word in each sentence. However, as we show in Section \S\ref{subsection:semeval} we still obtain sufficiently high WSI results.

The steps for performing Scalable WSI are summarized in Fig. \ref{fig:flow}. We elaborate on each step below, using English Wikipedia as a running example.\footnote{The Wikipedia corpus is based on a dump from August 2020, with text extracted using WikiExtractor \cite{Wikiextractor2015}.}

\textbf{Annotation:} We run \textit{BERT-large-cased-whole-word-masking} on English Wikipedia, inferring substitutes for all corpus positions. For positions that correspond to single-token words,\footnote{We exclude single-character tokens, stopwords and punctuation.} we consider the predicted words, filter stop-words, lemmatize the remaining words \cite{spacy}, and store the top-5 most probable lemmas to disk. This step takes 5 hours on 20 cloud-based GPU machines (total of 100 GPU hours), resulting in 1.63B tokens with their corresponding top-5 lemmas.

\textbf{Inverted Word Index:} We create an inverted index mapping from each single-token word to its corpus occurrences (and their corresponding top-5 lemmas). This takes 5 minutes on a 96 cores CPU machine, and 10GB of disk. 

\textbf{Sense Induction:}
For each of 16,081 lemmas corresponding to single-token words, we retrieve random 1000 instances,\footnote{The clustering algorithm scales super-linearly with the number of instances. To reduce computation cost for tokens that appear more than 1000 times in the dataset, we sample $\min($numOccur,$1000)$ instances for each token word, and cluster given the subset of instances. We then associate each of the remaining instances to one of the clusters as explained in the final step of the algorithm.} and induce senses using the community-based algorithm described in \S\ref{subsection:community_detection}. This process requires 30 minutes on the 96-core CPU machine, and uses 100MB of disk space. The average number of senses per lemma is 3.13. Each sense is associated with up to 100 representative words, which represent the highest-degree words in the sense's community. 
Table \ref{table:bass_communities} shows the 5 senses found for the word \emph{bass} with their top-5 representative words. See additional examples in Fig. \ref{fig:examples} and Appendix \ref{appendix:more_examples}.

\begin{table}[t]
\small
\centering
\resizebox{0.8\linewidth}{!}{%
\begin{tabular}{lllll}
\textbf{bass$_0$}   & \textbf{bass$_1$} & \textbf{bass$_2$} & \textbf{bass$_3$}  & \textbf{bass$_4$} \\
bassist    & double   & fish     & tenor     & trap    \\
guitar     & second   & bottom   & baritone  & swing   \\
lead       & tail     & perch    & voice     & heavy   \\
drum       & steel    & shark    & soprano   & dub     \\
rhythm     & electric & add      & singer    & dance   \\
\end{tabular}
}
\caption{Top 5 representatives of the sense-specific communities of word \textit{bass}. The communities roughly match to bass as a musical instrument, register, fish species, voice and in the context of Drum\&Bass}
\label{table:bass_communities}
\vspace{-4mm}
\end{table}

\textbf{Tagging:} 
Each of the remaining word-occurrences is associated with a sense cluster by computing the Jaccard similarity between the occurrences' top-5 lemmas and the cluster representatives, and choosing the cluster that maximizes this score. For example, an occurrence of the word \emph{bass} with lemmas \emph{tenor, baritone, lead, opera, soprano} will be associated with bass$_3$.
This takes 100 minutes on  96-core machine, and 25GB of storage.

\section{Sense Clustering Algorithm}
\label{subsection:community_detection}
We replace the hierarchical clustering algorithm used by \citet{amrami2018word,amrami2019towards} with a community-detection, graph-based clustering algorithm. 
One major benefit of the community detection algorithms is that they naturally produces a dynamic number of clusters, and provide a list of interpretable discrete representative lemmas for each cluster.
We additionally found this method to be more stable. 

Graph-based clustering for word-sense induction
typically constructs a graph from word occurrences or collocations, where the goal is to identify sense-specific sub-graphs within the graph that best induce different senses \cite{klapaftis2008word, klapaftis2010word}. We instead construct the graph based on word substitutes. Following \citet{jurgens2011word}, we pose identifying sense-specific clusters as a \emph{community detection problem}, where a community is defined as a group of connected nodes that are more connected to each other than to the rest of the graph.

\paragraph{Graph construction}
For each word $w$ in the vocabulary, we construct a graph $G_w=(V_w,E_w)$ where each vertex $v\in V_w$ is a substitute-word predicted by the MLM for $w$, and an edge $(u, v) \in E_w$ connects substitutes that are predicted for the same instance. The edge is weighted by the number of instances in which both $u$ and $v$ were predicted.
More formally, let $X = \{x_w^i\}_{i=1}^n$ bet the set of all top-$k$ substitutes for $n$ instances of word $w$, and $x_w^i = \{{w'}^j_{x_w^i}\}_{j=1}^k$ represents the $k$ top substitutes for the $i$th instance of word $w$. 
The graph $G_w$ is defined as follows:
\begin{equation*}
\begin{split}
V_w &=\{u: \exists i~u \in x_w^i \} \\
E_w &=\{(u, v) : \exists i~u \in x_w^i \wedge  v \in x_w^i\} \\
W(u, v) &= |\{ i : (u, v) \in x_w^i\}|
\end{split}
\end{equation*}
\noindent

\paragraph{Community detection} A community in a sub-graph corresponds to a set of tokens that tend to co-occur in top-$k$ substitutes of many instances, and not co-occur with top-$k$ substitutes of other instances. This corresponds well to senses and we take community's nodes as sense's representatives. 

We identify communities using the fast \textit{``Louvain''} method \cite{blondel2008fast}. Briefly, Louvain searches for an assignment of nodes to clusters such that the \emph{modularity score} $Q$---which measures the density of edges inside communities compared to edges between communities---is maximized:
\begin{equation*}
Q = \frac{1}{2m}\sum_{u~v} \left[W(u, v)- \frac{k_u k_v}{2m} \right]\delta(c_u, c_v) \\
\end{equation*}
\noindent $m$ is the sum of all edge weights in the graph, $k_u=\sum_v W(u,v)$ is the sum of the weights of the edges attached to node $u$, $c_u$ is the community to which $u$ is assigned, and $\delta$ is Kronecker delta function. This objective is optimized using an iterative heuristic process.
For details, see \citet{blondel2008fast}.

\section{Intrinsic Evaluation of Clustering Algorithm}
\label{section:evaluation}

We start by \emph{intrinsically evaluating} the WSI clustering method on: (a) \semeval2010 and \semeval2013; and (b) a new test set we develop for large-scale WSI. In section \ref{section:embs_eval}, we additionally \textit{extrinsically evaluate} the accuracy of static embeddings derived from a sense-induced Wikipedia dataset.

When collecting word-substitutes, we lemmatize the top-k list, join equivalent lemmas, remove stopwords and the target word from the list, and keep the top-5 remaining lemmas.

\subsection{SemEval Evaluation}
\label{subsection:semeval}
We evaluate the community-based WSI algorithm on two WSI datasets:
\semeval 2010 Task 14 \cite{manandhar2010semeval} and \semeval 2013 Task 13 \cite{jurgens2013semeval}. Table \ref{table:semeval} compares our method to \citet{amrami2018word, amrami2019towards} and AutoSense \cite{amplayo2019autosense}, which is the second-best available WSI method.
Bert-noDP/DP are taken from \citet{amrami2019towards}. Bert-DP uses ``dynamic patterns'' which precludes wide-scale application. We follow previous work
\cite{manandhar2010semeval, komninos2016structured, amrami2019towards} and evaluate \semeval 2010 using F-Score and V-Measure and \semeval 2013 using Fuzzy Normalized Mutual Information (FNMI) and Fuzzy
B-Cubed (FBC) as well as their geometric mean (AVG).
Our method performs best on \semeval 2010 and comparable to state-of-the-art results on \semeval 2013. The algorithm performs on-par with the Bert-noDP method, and does not fall far behind the Bert-DP method.
We now turn to assess the end-to-end induction and tagging over Wikipedia.

\subsection{Large Scale Manual Evaluation}
\label{section:tagging_evaluation}

We evaluate our method on large corpora by randomly sampling 2000 instances from the sense-induced Wikipedia, focusing on frequent words with many senses. We manually annotate the samples' senses without access to the automatically induced senses, and then compare our annotations to the system's sense assignments. We publicly release our manual sense annotations.

\begin{table}[t]
\small
\centering
\resizebox{\linewidth}{!}{%
\begin{tabular}{l|lll}
\textbf{Model}               & \textbf{F-S}       & \textbf{V-M}     & \textbf{AVG} \\
\hline 
AutoSense           & 61.7         & 9.8          & 24.59        \\
Bert-noDP & 70.9 \small{(0.4)}   & 37.8 \small{(1.5)}   & 51.7 \small{(1.2)}   \\
Ours & \textbf{70.95 \small{(0.63)}} & \textbf{40.79 \small{(0.19)}} & \textbf{53.79 \small{(0.31)}} \\
\hline
Bert-DP & 71.3 \small{(0.1)}   & 40.4 \small{(1.8)}   & 53.6 \small{(1.2)}
\end{tabular}}
\\~\\~\\
\resizebox{\linewidth}{!}{%
\begin{tabular}{l|lll}
\textbf{Model}               & \textbf{FNMI}       & \textbf{FBC}     & \textbf{AVG} \\
\hline 
AutoSense           & 7.96         & 61.7          & 22.16        \\
Bert-noDP & 19.3 \small{(0.7)}   & \textbf{63.6 \small{(0.2)}}   & \textbf{35.1 \small{(0.6)}}   \\
Ours & \textbf{19.42 \small{(0.39)}} & 61.98 \small{(0.12)} & 34.69 \small{(0.33)} \\
\hline
Bert-DP & 21.4 \small{(0.5)}   & 64.0 \small{(0.5)}   & 37.0 \small{(0.5)}\\
\end{tabular}}
\caption{Evaluation on the \semeval2010 (top) and \semeval2013 (bottom) datasets. We report mean (STD) scores over 10 runs.}
\label{table:semeval}
\vspace{-4mm}
\end{table}

\paragraph{Sampling and Manual Annotation}
We used a list of 20 ambiguous words from \textit{CoarseWSD-20} \cite{loureiro2021analysis}. The full list and per-word results can be found in Appendix \ref{appendix:tagging_evaluation}. For each word we sampled 100 passages from English Wikipedia with the target word, including inflected forms (case insensitive).
Unlike \textit{CoarseWSD-20}, we sampled examples without any respect to a predefined set of senses. For example, the only two senses that appear in \textit{CoarseWSD-20} for the target word \textit{arm} are \textit{arm (anatomy)}, and \textit{arm (computing)}, leaving out instances matching senses reflecting \textit{weapons, subdivisions, mechanical arms} etc.

With the notion that word sense induction systems should be robust to different annotations schemes, we gave two fluent English speakers 100 sentences for each of the 20 ambiguous words from \textit{CoarseWSD-20}. Annotators were not given a sense inventory. Each annotator was asked to label each instance with the matching sense \textit{according to their judgment}. For example, for the target word \textit{apple} in the sentence \textit{``The iPhone was announced by \textit{Apple} CEO."}, annotators can label the target sense with \textit{Apple Inc.}, \textit{Apple The Company} etc. Annotation Guidelines are available in Appendix \ref{appendix:annotation_guidelines}.

On average annotators labeled $6.65$ senses per word ($5.85$ and $7.45$ average clusters per word for the two annotators). This is more than the $2.65$ average senses according to \textit{CoarseWSD-20} and less than \wordnet's $9.85$.


\paragraph{Results}
We report our system's performance alongside two additional methods: A strong baseline of the most frequent sense (MFS), and Babelfy \cite{moro2014entity}---the sense disambiguation system used in \babelnet~(Tested using Babelfy live version April 2021). Differently from the latter, our system does not disambiguates but induces senses, therefore, clusters are not labeled with a sense tag from a sense inventory. Instead, we represent senses to annotators using a list of common substitute words and a few examples. Thus, after annotating the Wikipedia passages, we additionally asked annotators to name the system's clusters with the same naming convention as in their annotations.

Given a similar naming convention between systems and annotators, we report F1 scores of systems' tagging accuracy with respect to the manual annotations. We report F1 averaged over words in Table \ref{table:tagging_evaluation}. Our system outperforms both baselines, despite Babelfy having access to a list of predefined word senses. A full by-word table and comprehensive results analysis are in Appendix \ref{appendix:tagging_evaluation}.

While a 1-to-1 mapping between system clusters and manual senses is optimal, our system sometimes splits senses into smaller clusters, thus annotators will name two system clusters with the same label. Therefore it is also important to report the number of clusters produced by the system comparing to the number of senses after the annotators merged similar clusters. Our system produced $7.25$ clusters with $2.25$ clusters on average merged by the annotators.\footnote{This is partially due to using clusters from two casing (\eg \textit{bank} and \textit{Bank}), some of the merges share sense meaning but of different casing.}
Additionally, in rare cases our system encapsulates a few senses in a single cluster: this happened 3 and 5 times for both annotators across all the dataset. 

\begin{table}[t]
\small
\centering
\begin{tabular}{c|ccc}
        & \textbf{MFS}   & \textbf{Babelfy} & \textbf{Ours}  \\
\hline
Ann \#1 & 49.55 & 41.5    & \textbf{89.05} \\
Ann \#2 & 49.9    & 41.95      & \textbf{85.95} \\
\hline
average & 49.72    & 41.72      & \textbf{87.50}   
\end{tabular}
\caption{Classification F1 scores for MFS, Babelfy and our proposed system by annotator on our manually annotated dataset.}
\label{table:tagging_evaluation}
\vspace{-4mm}
\end{table}

\section{Application to Scientific Corpora}
\label{section:biomedical_data}

A benefit of a WSI approach compared to WSD methods is that it does not rely on a pre-specified sense inventory, and can be applied to any corpus for which a BERT-like model is available.
Thus, in addition to the Wikipedia dataset that has been presented throughout the paper, we also automatically induce senses over a corpus of 31 million PubMed Abstracts,\footnote{\href{https://www.nlm.nih.gov/databases/download/pubmed\_medline.html}{www.nlm.nih.gov/databases/download/pubmed\_medline}} using SciBERT \cite{Beltagy2019SciBERT}. As this dataset is larger than the Wikipedia dump, the process required roughly 145 GPU hours and resulting in $14,225$ sense-annotated lemmas, with an average number of $2.89$ senses per lemma.

This dataset highlights the data-driven advantages of sense-induction: the algorithm recovers many senses that are science specific and are not represented in the Wikipedia corpora. While performing a wide-scale evaluation of the scientific WSI is beyond our scope in this work, we do show a few examples to qualitatively demonstrate the kinds of induced senses we get for scientific texts.

For each of the words \emph{mosaic, race} and \emph{swine} we show the induced clusters and the top-5 cluster representatives for each cluster.

\vspace{0.5em}
\begin{center}
\small
\centering
\begin{tabular}{llllll}
\textbf{mosaic$_{0}$} & \textbf{mosaic$_{1}$} & \textbf{mosaic$_{2} $}& \textbf{mosaic$_{3}$} \\
virus & partial & mixture & mixed\\
dwarf & chimeric & landscape & genetic\\
mild & congenital & combination & spatial\\
cmv & heterozygous & pattern & functional\\
stripe & mutant & matrix & cellular\\
\end{tabular}
\end{center}
\vspace{0.5em}

While senses mosaic$_{0}$ (the common mosaic virus of plants) and mosaic$_{2}$ (``something resembling a mosaic", ``mosaic of..") are represented in Wikipedia, senses mosaic$_1$ (the mosaic genetic disorder) and mosaic$_3$ (mosaic is a quality, e.g., ``mosaic border'', ``mosaic pattern'') are specific to the scientific corpora (The Wikipedia corpora, on the other hand, includes a sense of mosaic as a decorative art-form, which is not represented in Pubmed).

\vspace{0.5em}
\resizebox{\linewidth}{!}{%
\small
\begin{tabular}{llll}
\textbf{race$_{0}$} & \textbf{race$_{1}$} & \textbf{race$_{2}$} & \textbf{race$_{3}$} \\
racial & exercise & class & pcr \\
ethnicity & run & group & clone \\
black & training & state & sequence \\
rac & competition & population & rt \\
gender & sport & genotype & ra \\
\end{tabular}
}
\vspace{0.5em}


\noindent Senses race$_{0}$ (ethnic group), race$_{1}$ (competition) and race$_{2}$ (population/civilization) are shared with wikipedia, while the sense race$_3$ (``Rapid amplification of cDNA ends'', a technique for obtaining the sequence length of an RNA transcript using reverse transcription (RT) and PCR) is Pubmed-specific. 

\vspace{0.5em}
\resizebox{0.8\linewidth}{!}{%
\begin{tabular}{lll}
\textbf{swine$_{0}$} & \textbf{swine$_{1}$} & \textbf{swine$_{2}$}\\
pig & seasonal & patient\\
porcine & avian & infant\\
animal & influenza & group\\
livestock & pandemic & case\\
goat & bird & myocardium\\
\end{tabular}
}
\vspace{0.5em}

\noindent Here swine$_{1}$ captures the Swine Influenza pandemic, while swine$_{2}$ refers to swine as experimental Pigs.

\section{Sense-aware Information Retrieval}
\label{section:senseful_ir}
\begin{figure*}[t!]
\centering
\includegraphics[width=\linewidth]{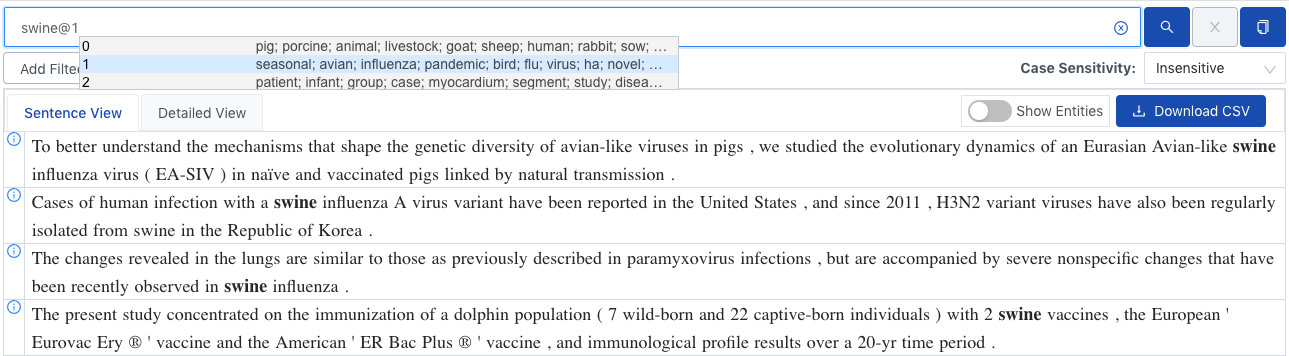}
\caption{User interaction in SPIKE when looking for the word \textit{``swine"} in its \textit{``swine flu"} sense. (Unlike the animal/experimental pig senses)}
\label{fig:swine_in_spike}
\vspace{-4mm}
\end{figure*}

An immediate application of a high quality sense-tagged corpus is sense-aware retrieval.
We incorporate the sense information in the SPIKE extractive search system \cite{Shlain2020SyntacticSB}\footnote{\href{https://spike.apps.allenai.org}{spike.apps.allenai.org}} for Wikipedia and Pubmed datasets.
When entering a search term, suffixing it with @ triggers sense selection allowing to narrow the search for the specific sense.
Consider a scientist looking for PubMed occurrences of the word  \textit{``swine"} in its influenza meaning. As shown in Figure \ref{fig:swine_in_spike}, this can be easily done by writing ``swine@'' and choosing the second item in the resulting popup window. The outputs are sentences with the word \textit{``swine"} in the matching sense. As far as we know, SPIKE is the first system with such WSI capabilities for IR. Similarly, \citet{blloshmi2021ir} suggested to enhance IR with sense information, but differently from us, this is done by automatically tagging words with senses from a predefined inventory.

\section{Sense-aware Static Embeddings}
\label{section:senseful_w2v}

Learning static word embeddings of sense-ambiguous words is a long standing research goal \cite{reisinger2010multi, huang2012improving}.
There are numerous real-world tasks where context is not available, precluding the use of contextualized-embeddings. These include Outlier Detection \cite{camacho2016find, blair2016automated}, Term Set Expansion \cite{roark2000noun} the Hypernymy task \cite{breit-etal-2021-wic}, etc.
Additionally, static embeddings are substantially more efficient to use, can accommodate larger vocabulary sizes, and can accommodate efficient indexing and retrieval.
Yet, despite their flexibility and success, common word embedding methods still represent ambiguous words as a single vector, and suffer from the inability to distinguish between different meanings of a word \cite{camacho2018word}.

Using our sense-tagged corpus we suggest a simple and effective method for deriving sense-aware static embeddings: We run an off-the-shelf embedding algorithm,\footnote{We use the CBOW variant of the word2vec algorithm \cite{mikolov2013distributed} as implemented in Gensim \cite{rehurek_lrec}. We derive 100-dimensional embeddings using the negative-sampling algorithm and a window size of 5.}
on the corpus where single-token words are replaced with a concatenation of the word and its induced sense (\eg \textit{``I caught a bass."} becomes \textit{``I caught@0 a bass@2."}). This makes the embedding algorithm learn embeddings for all senses of each word out-of-the-box.\footnote{A similar approach was used by \citet{iacobacci-etal-2015-sensembed} over a corpus which was labeled with BabelNet and WordNet senses.} 
An integral property of the embedding algorithm is that it represents both the sense-annotated tokens and the other vocabulary items in the same embedding space --- this helps sense inferring about words that are represented in the MLM as multi-tokens words (Even though these correspond to less-frequent and often less ambiguous words \cite{hernandez2016testing, fenk2010frequency, zipf1945meaning}). For example, in the top-5 nearest neighbours for the different \emph{bass} senses as shown below, \textit{smallmouth} and \textit{pumpkinseed}, multi-token words in \textsc{BERTlarge}'s vocabulary, are close neighbours the \textit{bass} instances that correspond to the \textit{fish} sense.
\vspace{-5mm}
\begin{center}
\resizebox{\linewidth}{!}{%
\begin{tabular}{lllll}
\textbf{bass$_0$} & \textbf{bass$_1$} & \textbf{bass$_2$} & \textbf{bass$_3$} & \textbf{bass$_4$} \\
guitar$_0$ & tuba & crappie & baritone$_0$ & synth \\
drums$_0$ & trombone$_0$ & smallmouth & tenor$_0$ & drum$_1$ \\
guitar$_3$ & horn$_0$ & pumpkinseed & alto$_0$ & synths \\
keyboards$_0$ & flute$_0$ & sunfish & bassoon & breakbeats \\
keyboard$_0$ & trumpet$_0$ & perch$_0$ & flute$_0$ & trap$_4$ \\
\end{tabular}
}
\end{center}
Note that some neighbours are sense annotated (single-token words that were tagged by our system), while others are not (multi-token words).

For English Wikipedia, we obtain a total vocabulary of 1.4M forms, $90,023$ of which are sense-annotated. Compared to the community-based representative words, the top neighbours in the embedding space tend to capture members of the same semantic class rather than direct potential replacements.

\section{Sense-aware Embeddings Evaluation}
\label{section:embs_eval}

\subsection{WiC Evaluation}
\label{section:wic}

\citet{pilehvar-camacho-collados-2019-wic} introduced the \wic~dataset for the task of classifying word meaning in context. Each instance in \wic~has a target word and two contexts in which it appears. The goal is to classify whether the word in the different contexts share the same meaning. \eg given two contexts: \textit{There's a lot of trash on the \underline{bed} of the river} and \textit{I keep a glass of water next to my \underline{bed} when I sleep}, our method should return \textit{False} as the sense of the target word \textit{bed} is different.

Our method is the following: Given the sense-aware embeddings, a target word $w$ and two contexts, we calculate the context vector as the average of the context words. The matching sense vector is the closest out of all $w$ embeddings. We then classify the contexts as corresponding to the same meaning if the cosine distance of the found sense embedding is more than threshold apart. We do not use the train set. The threshold is optimized over the development set and fixed to $0.68$.

\begin{table}[t]
\small
\centering
\begin{tabular}{l|c}
\textbf{Method}   & \textbf{Acc.}  \\
\hline
JBT \cite{pelevina2017making}    & 53.6  \\
\textbf{Sense-aware Embeddings (this work)}  & \textbf{58.3} \\
\hline
SW2V* \cite{mancini-etal-2017-embedding}  & 58.1  \\
DeConf* \cite{pilehvar2016conflated} & 58.7  \\
\textbf{LessLex*} \cite{colla2020lesslex} & \textbf{59.2} \\
\end{tabular}
\caption{Accuracy scores on the \wic~dataset. Systems marked with * make use of external lexical resources.}
\label{table:wic_results}
\vspace{-4mm}
\end{table}

This task has a few tracks, we compare our embeddings systems to the best performing methods from the \textit{Sense Representations} track. Of these, JBT \cite{pelevina2017making}, a lexical embedding method, is the only one that does not use an external lexical resource (induction). The results in Table \ref{table:wic_results} show accuracy on this task. We outperform the induction method, and are on-par with the lexicon-based methods, despite not using any external lexical resource.

\begin{table}[t]
\small
\centering
\begin{tabular}{l|cc}
\textbf{Word Embeddings}         & \textbf{OPP}    & \textbf{Acc.} \\
\hline
\glove                   & 93.31 & 65   \\
word2vec                     & 93.31   & 68   \\
DeConf & 93.37 & 73 \\
Ours (Skip-gram) & 96.31 & 83.5 \\
\textbf{Ours (CBOW)}      & \textbf{96.68}  & \textbf{86} \\
\end{tabular}
\caption{OPP and Accuracy on the \textit{25-7-1-8} dataset.}
\label{table:outlier_results}
\vspace{-4mm}
\end{table}

\subsection{Evaluation via Outlier Detection}
\label{subsection:outlier_detection}
Another setup for evaluating word embeddings is that of \emph{outlier detection}: given a set of words, identify which one does not belong to the set \cite{blair2016automated}. Outlier detection instances are composed of in-group elements and a set of outliers from a related semantic space. In each evaluation round, one outlier is added to the in-group items, and the algorithm is tasked with finding the outlier.
Existing outlier detection datasets either did not explicitly target sense-ambiguous words (\textit{8-8-8} \cite{camacho2016find},  \textit{WikiSem500} \cite{blair2016automated}) or explicitly \emph{removed} ambiguous words altogether (\textit{25-8-8-sem} \cite{andersen2020one}). 

\paragraph{Ambiguity-driven Outlier Detection.}  We construct a challenge set for outlier detection that specifically targets ambiguous cases.
In order to account for sense ambiguity, we add a \emph{distractor} to each of the in-group sets: the distractor is an item which has multiple senses, where the most salient sense does not belong to the group, while another sense does belong to the group. For example:

\textbf{In-group:} \textit{zeus, hades, poseidon, aphrodite, ares, athena, artemis}\\
\textbf{Outliers:} \textit{mercury, odysseus, jesus, sparta, delphi, rome, wrath, atlanta}\\
\textbf{Distractor:} \textit{nike}

Here, a model which does not explicitly represent the greek-god sense of \emph{nike} is likely to place it far away from the in-group instances, causing it to be mistakenly marked as the outlier.

The starting point for our dataset is \textit{25-8-8-Sem} \cite{andersen2020one}. This dataset contains 25 test groups, each with 8 in-group elements and 8 outliers, resulting in 200 unique test cases. The outliers are sorted in a decreasing degree of relatedness to the in-group elements. In our dataset we replace one of the in-group elements with an ambiguous distractor. For example, in the Greek-gods case above, we replaced the original 8\textsuperscript{th} item (\textit{``hera"}) with the ambiguous distractor \emph{nike}.
\footnote{We additionally changed terms that are debatably ambiguous and changed the \textit{``African animals"} group to the more general \textit{``animals"} as no distractors were found.}
The dataset consists of 25 groups of 7 non ambiguous group elements, 1 distractor and 8 outliers (\ourds), similarly resulting 200 unique test cases.

\noindent\textbf{Method}$\,\,\,$ Following \citet{camacho2016find}, we rank each word likelihood of being the outlier by the average of all pair-wise semantic similarities of the words in $W\backslash \{w\}$. Therefore if $w$ is an outlier, this score should be low. See Appendix \ref{appendix:outlier_detection_method} for additional details.

\noindent\textbf{Metrics}$\,\,\,$
\citet{camacho2016find} proposed evaluating outlier detection using the accuracy (The fraction of correctly classified outliers among the total cases) and Outlier Position Percentage (OPP) metric. OPP indicates how close outliers are to being classified correctly:
\begin{equation*}
OPP = \frac{\sum_{W \in D} \frac{OP(W)}{\vert W \vert -1}}{\vert D \vert} \times 100
\end{equation*}

\noindent where $OP(W)$ is the position of the outlier according to the algorithm.

\noindent\textbf{Results}$\,\,\,$
In Table \ref{table:outlier_results} we report performance of on the \ourds~set. Word2vec and \glove~accuracy scores are low while having high OPP scores. This is the expected behaviour for embeddings without sense awareness. These will position the distractor and the outlier furthest away from the group items while not designed to make the hard decision required for high Accuracy. Our sense-aware embeddings strongly outperform GloVe and word2vec which do not include senses.
Our embeddings also outperform the word embeddings proposed in DeConf \cite{pilehvar2016conflated}, which are the best performing sense embeddings on WiC which are also publicly available.

\section{Conclusion}
\label{section:conclusion}
We show that substitution-based word-sense induction algorithms based on word-substitutions derived from MLMs are easily scalable to large corpora and vocabulary sizes, allowing to efficiently obtain high-quality sense annotated corpora. 
We demonstrate the utility of such large-scale sense annotation, both in the context of a scientific search application, and for deriving high-quality sense-aware static word embeddings. 

As a secondary contribution, we also develop a new variant of the Outlier Detection evaluation task, which explicitly targets ambiguous words.

\section{Acknowledgments}
This project has received funding from the European Research Council (ERC) under the European Union’s Horizon 2020 research and innovation programme, grant agreement No. $802774$ (iEXTRACT).

\bibliography{anthology,custom}
\bibliographystyle{acl_natbib}

\clearpage
\appendix
\section{Additional Examples}
\label{appendix:more_examples}
Due to limit of space we provide additional examples in the appendix. We start with the senses found for the word \textit{face}:

\vspace{0.5em}

\noindent
{\small
\begin{tabular}{llll}
\multicolumn{4}{l}{\textbf{Representatives}} \\
\textbf{face$_0$} & \textbf{face$_1$} & \textbf{face$_2$} & \textbf{face$_3$} \\
confront & head & look & side \\
meet & front & address & line \\
encounter & name & point & wall \\
suffer & cheek & serve & surface \\
experience & body & toward & slope \\
\end{tabular}}
\\

\noindent
{\small
\begin{tabular}{llll}
\multicolumn{4}{l}{\textbf{Neighbours}} \\
\textbf{face$_0$} & \textbf{face$_1$} & \textbf{face$_2$} & \textbf{face$_3$} \\
meet$_3$ & hand$_0$ & faced$_2$ & slope$_0$ \\
challenge$_3$ & forehead$_0$ & sit$_1$ & rim$_0$ \\
suffer$_0$ & hands$_0$ & hang$_1$ & flank$_2$ \\
confront$_0$ & nose$_0$ & facing$_2$ & ridge$_4$ \\
lose$_1$ & eyes$_3$ & rotate$_0$ & slope$_1$ \\
\end{tabular}}
\vspace{0.5em}

The face senses refer to meeting/confronting, the body part, turn/look and side, respectively.

Here we present two senses of the word \textit{orange}, corresponding to the color and fruit:

\vspace{0.5em}

\noindent
\resizebox{0.8\linewidth}{!}{%
\begin{tabular}{llll}
\multicolumn{2}{l}{\textbf{Representatives}} & \multicolumn{2}{l}{\textbf{Neighbours}} \\
\textbf{orange$_0$} & \textbf{orange$_1$} & \textbf{orange$_0$} & \textbf{orange$_1$} \\
yellow & apple & yellow$_0$ & apple$_0$\\
red & lemon & purple$_0$ & avocado \\
amber & lime & amber$_0$ & almond \\
pink & fruit & blue$_0$ & apple$_1$ \\
olive & banana & orangish & apricot \\
\end{tabular}
}

\vspace{0.5em}
Finally we present the senses for \textit{Jordan}:
\vspace{0.5em}

\noindent
{\small
\begin{tabular}{llll}
\multicolumn{4}{l}{\textbf{Representatives}}\\
\textbf{Jordan$_0$} & \textbf{Jordan$_1$} & \textbf{Jordan$_2$} & \textbf{Jordan$_3$} \\
Johnson & Jerusalem & David & River \\
Jones & Palestine & Jason & Zion \\
Jackson & Israel & Joel & Water \\
Murray & Yemen & Justin & City \\
Mason & Turkey & Jonathan & water \\
\end{tabular}}
\\ 

\noindent
{\small
\begin{tabular}{llll}
\multicolumn{4}{l}{\textbf{Neighbours}}\\
\textbf{Jordan$_0$} & \textbf{Jordan$_1$} & \textbf{Jordan$_2$} & \textbf{Jordan$_3$} \\
Jones$_1$ & Kuwait$_1$ & Jeremy$_1$ & Huleh \\
Kramer$_1$ & Lebanon$_0$ & Aaron$_0$ & Yarkon \\
Allen$_0$ & Syria$_0$ & Justin$_0$ & Arabah \\
Mack$_0$ & Iraq$_0$ & Brandon$_0$ & Khabur \\
Robinson$_0$ & Sudan$_1$ & Josh$_0$ & Tyropoeon \\
\end{tabular}}

\vspace{1ex}
Here the clusters correspond to Jordan the surname, the country, first name and the Jordan River, respectively.

\section{Annotation Guidelines for Manual Evaluation}
\label{appendix:annotation_guidelines}

The objective of this task is to annotate word-meanings of 20 ambiguous words in a total of 2000 different contexts.

What is word-meaning? Words have different meanings in different contexts, for example, in the sentence: \textit{``there is a \underline{light} that never goes out"}, the word \textit{``light"} refers to any device serving as a source of illumination. While \textit{``light"} in the sentence \textit{``\underline{light} as a feather"} refers to the comparatively little physical weight or density of an object.

\textbf{Step 1:}

In this dataset we examine 20 ambiguous words as targets. For each of these words we collected 100 sentences in which the target word appears. For every sentence in the 100 set per target word, you will be asked to write a short label expressing the meaning of the target word in that particular context.

For example, here are three sentences with the target word \textit{``light"}, each with its possible annotation.

1. \textit{``there is a \underline{light} that never goes out"} $\rightarrow$ visible light.

2. \textit{``\underline{light} as a feather"} $\rightarrow$ light as in weight.

3. \textit{``magnesium is a \underline{light} metal"} $\rightarrow$ light as in weight.

Note that in this example the annotator found the second and third meanings of the word \textit{``light"} to be the same and therefore labeled them with the same label.\footnote{For ease of use for future evaluators, at the end of this step, both annotators picked a single naming convention when two labels referred to the same sense. Names of labels that were used only by one annotator were not changed.}

While some annotations are indeed intuitive, labeling word-meanings when the target word is part of a name can be challenging. Here are a few guidelines for such use case:

Whenever a target word appeared as part of a name (Person, Organization etc.), one of three heuristics should be used\footnote{Some of the dissimilarities between the annotations are with respect the tension between the second and third guidelines.}:

1. If the target word is the surname of a person, the example should be tagged \textit{surname}.\footnote{As opposed to Babelfy, there was no attempt for entity linking, so all persons were tagged the same.}

2. If the entity (as a whole) refers to one of the word-meanings, it should be labeled as such. For example, \textit{Quitobaquito Springs} label should refer to a natural source of water.

3. If the target word is part of a name different from the original word-meaning, it should be tagged as \textit{Part of Name}. This includes song names, companies (\textit{Cold Spring Ice}), restaurants etc. Possible exceptions for this case are when a specific named entity is significantly frequent.

\textbf{Step 2:} \footnote{This step is presented to annotators once step 1 is done for all words}

For each of the target words you labeled, you will now receive a short list of indirect word-meaning definitions. Indirect word-meanings are composed of:

(a) A list of 10 words that may appear instead of the target word in specific contexts

(b) A list of 5 sentences in which the target word has this specific word-meaning.

For example, this is a possible indirect word-meaning for the target word \textit{``Apple"}, representing the fruit, as opposed to the tech company:

\textbf{Alternatives:}
orange, olive, cherry, lime, banana, emerald, lemon, tomato, oak, arrow, 

\textbf{Sentences in which Apple appears in this word-meaning:} 

\textit{``He and his new bride planted \underline{apple} trees to celebrate their marriage."}

\textit{``While visiting, Luther offers Alice an \underline{apple}."}

\textit{``When she picks the \underline{apple} up, it is revealed that Luther has stolen a swipe card and given it to Alice to help her escape."}

You will be asked to label the indirect word-meanings with one of the labels you used in step 1. If no label matches the indirect word-meaning you are allowed to propose a new label or define it to be “Unknown". Additionally, if you find several indirect word-meanings too close in meaning, label them the same.

\section{Analysis of Manual Evaluation}
\label{appendix:tagging_evaluation}

\begin{table*}[t]
\centering
\begin{tabular}{lccccc|ccccc}
& \multicolumn{5}{c}{Annotator \#1} & \multicolumn{5}{c}{Annotator \#2} \\
Word &  MFS &  Babelfy &  Ours &  F2R &  Ent. &  MFS &  Babelfy &  Ours &  F2R &  Ent. \\
\hline
Apple  & 48  & 69  & \textbf{94}  & 0.92  & 0.71  & 47  & 66  & \textbf{86}  & 0.89  & 0.05  \\
Arm  & 34  & 31  & \textbf{89}  & 0.52  & 0.87  & 34  & 33  & \textbf{85}  & 0.52  & 0.83  \\
Bank  & 48  & 61  & \textbf{94}  & 0.92  & 0.78  & 46  & 61  & \textbf{85}  & 0.85  & 0.69  \\
Bass  & 61  & 6  & \textbf{82}  & 1.56  & 0.64  & 65  & 17  & \textbf{83}  & 1.86  & 0.62  \\
Bow  & 31  & 14  & \textbf{80}  & 0.45  & 0.80  & 32  & 16  & \textbf{80}  & 0.47  & 0.83  \\
Chair  & 66  & 29  & \textbf{90}  & 1.94  & 0.66  & 67  & 31  & \textbf{86}  & 2.03  & 0.63  \\
Club  & 49  & 45  & \textbf{80}  & 0.96  & 0.78  & 53  & 50  & \textbf{77}  & 1.13  & 0.72  \\
Crane  & 39  & 36  & \textbf{86}  & 0.64  & 0.90  & 39  & 35  & \textbf{83}  & 0.64  & 0.69  \\
Deck  & 45  & 49  & \textbf{72}  & 0.82  & 0.80  & 48  & 52  & \textbf{71}  & 0.92  & 0.68  \\
Digit  & 87  & 96  & \textbf{99}  & 6.69  & 0.56  & 87  & 96  & \textbf{98}  & 6.69  & 0.38  \\
Hood  & 27  & 6  & \textbf{82}  & 0.37  & 0.88  & 28  & 5  & \textbf{82}  & 0.39  & 0.83  \\
Java  & 63  & 32  & \textbf{98}  & 1.70  & 0.67  & 63  & 31  & \textbf{97}  & 1.70  & 0.69  \\
Mole  & 37  & 32  & \textbf{90}  & 0.59  & 0.81  & 39  & 32  & \textbf{88}  & 0.64  & 0.73  \\
Pitcher & 95  & \textbf{97}  & \textbf{97}  & 19.00  & 0.20  & 95  & \textbf{97}  & \textbf{97}  & 19.00  & 0.20  \\
Pound  & 46  & 58  & \textbf{91}  & 0.85  & 0.75  & 46  & 58  & \textbf{91}  & 0.85  & 0.72  \\
Seal  & 30  & 48  & \textbf{88}  & 0.43  & 0.91  & 27  & 40  & \textbf{74}  & 0.37  & 0.80  \\
Spring  & 57  & 0  & \textbf{90}  & 1.33  & 0.63  & 56  & 0  & \textbf{88}  & 1.27  & 0.64  \\
Square  & 37  & 15  & \textbf{88}  & 0.59  & 0.86  & 36  & 15  & \textbf{85}  & 0.56  & 0.82  \\
Trunk  & 33  & 46  & \textbf{98}  & 0.49  & 0.90  & 33  & 46  & \textbf{92}  & 0.49  & 0.86  \\
Yard  & 58  & 60  & \textbf{93}  & 1.38  & 0.63  & 57  & 58  & \textbf{91}  & 1.33  & 0.59  \\
\hline
Average & 49.55  & 41.5  & \textbf{89.05}  & 2.11  & 0.74  & 49.9  & 41.95  & \textbf{85.95}  & 2.13  & 0.65  
\end{tabular}
\caption{Manually annotated set scores by annotator. The first three columns for each annotator reflect disambiguation and induction scores with respect to the most frequent sense, Babelfy and our proposed system. We also report F2R and normalized entropy (Ent).}
\label{table:tagging_evaluation_by_word}
\end{table*}

In table \ref{table:tagging_evaluation_by_word} we report a by-word analysis of our manual evaluation results. For each word we detail F1 scores of the most frequent sense (MFS), Babelfy, and our proposed system. Similarly to \citet{loureiro2021analysis}, we report the ratio of the first sense with respect to the rest (F2R) and normalized entropy\footnote{
Computed as 
$\frac{-\sum_{i=1}^{k} \frac{c_i}{n} \log \frac{c_i}{n}}{log(k)}$, where $k$ is the number of annotated senses, each of size $c_i$ and $n$ is the size of annotated examples per word, in our case $n=100$.
}
to reflect sense balance. All of which are reported per annotator.

\paragraph{Analysis}
Analysis of our system's error shows that for some words the system could not create a matching cluster for specific senses (to name a few examples, "yard" as a ship identifier and "impound/enclosure" sense for the word "pound"). It appears that a matching cluster was not created due to the low tally of these senses in the English Wikipedia, and indeed the two senses appeared only two and three times respectively in the 100 passages sample. Additionally, annotator 2 annotated in a more fine-grained manner that does not correspond to our system tendency to merge capitalized instances of the target word into a sense that corresponds to "part of named entity".

As described above, in rare cases our system merged two senses into a single cluster. For example, the same cluster of the word "trunk" contained occurrences which annotator 1 tagged either "human torso" or "tube-like organs" (like the pulmonary trunk). While such annotation was uncommon (3 out of 117 senses for annotator 1 and 5 out of 149 senses for annotator 2), it does affect our system's micro F1 score for the better. In case we do not allow such annotation our overall score drops from $87.52$ to $86.65$.


A comparison between Babelfy and our gold annotation shows a common mistake in its labeling where Babelfy attributes the vast majority of sentences to the same non-salient sense. For example, Babelfy attributes 77 out of 100 instances of \textit{hood} to "An aggressive and violent young criminal" - a sense that was not found even once in the manual annotation. While in a number of cases Babelfy used finer-grained sysnset groups than in our annotations we took into account any senses that are a subset of our annotated senses. For examples, Babelfy's \textit{"United States writer who lived in Europe; strongly influenced the development of modern literature (1885-1972)"} synset was attribute any instances from the senses \textit{surname} that refer to the writer Ezra Pound.




\section{Outlier Detection Method}
\label{appendix:outlier_detection_method}
When using a single-prototype vector-space models, \citet{camacho2016find} proposed a procedure for detecting outliers based on semantic similarity using \textit{compactness score}:
\begin{equation*}
c(w) = \frac{1}{n^2 - n} \sum_{w_i \in W \setminus \{w\}} \sum_{\substack{w_j \in W \setminus \{w\}\\w_i \ne w_j}} sim(w_i, w_j)
\end{equation*}
Where $D$ is the entire dataset and $W$ is defined as $\{w_1, w_2, \cdots, w_n, w_{n+1}\}$ where w.l.o.g. $\{w_1, w_2, \cdots, w_n\}$ are the group elements (including the distractor) and $w_{n+1}$ is the outlier.
We use the same procedure with an additional nuance, we expanded the procedure to receive more than a single vector representation per word such that it will fit multi-prototype embeddings (\eg our embeddings and DeConf) and case sensitive embeddings (\eg word2vec). When given as set of words (like $W \setminus \{w\}$ when calculating $c(w)$) we first find the relevant sense for each element before inferring the outlier.
\citet{camacho2016find} suggested calculating $c(w)$ using the \textit{pseudo inverted compactness score}.

\end{document}